\documentclass{article}

\PassOptionsToPackage{numbers, compress}{natbib}
\usepackage[preprint]{neurips_2026}

\usepackage[utf8]{inputenc} % allow utf-8 input
\usepackage[T1]{fontenc}    % use 8-bit T1 fonts
\usepackage{hyperref}       % hyperlinks
\usepackage{url}            % simple URL typesetting
\usepackage{booktabs}       % professional-quality tables
\usepackage{amsfonts}       % blackboard math symbols
\usepackage{nicefrac}       % compact symbols for 1/2, etc.
\usepackage{microtype}      % microtypography
\usepackage{xcolor}         % colors
\usepackage{graphicx}
\usepackage{float}
\usepackage{placeins}
\usepackage{fancyvrb}
\usepackage{xcolor}
\usepackage{fvextra}
\DefineVerbatimEnvironment{PromptVerbatim}{Verbatim}{
  breaklines=true,
  breakanywhere=true,
  fontsize=\tiny,
  tabsize=2
}

\title{Automated Synthesis and Adversarial Validation of Executable Causal Research Pipelines}

\author{%
  Irena Girshovitz\\
  School of Biomedical Engineering\\
  Tel Aviv University\\
  Tel Aviv, Israel \\
  \texttt{girshovitz1@mail.tau.ac.il} \\
   \AND
   Dan Zeltzer \\
   School of Public Health \\
   University of California, Berkeley\\
   Berkeley, CA, USA\\
   \texttt{dzeltzer@berkeley.edu} \\
   \AND
   Ran Gilad-Bachrach \thanks{Corresponding author}\\
   School of Biomedical Engineering\\
   Tel Aviv University\\
   Tel Aviv, Israel \\
   \texttt{rgb@tauex.tau.ac.il} \\
}

\makeatletter
\if@preprint
  \RenewEnviron{ack}{}
\fi
\makeatother

\begin{document}

\maketitle

\begin{abstract}
  While automated research systems promise to accelerate empirical analysis, they are prone to silent failures: instances in which analysis code executes successfully yet relies on invalid causal assumptions. We present the Artificial Intelligence (AI)-based Epidemiology Research Assistant (ARA), a framework that makes these failures visible by explicitly encoding causal design principles, study-specific assumptions, and methodological constraints. ARA integrates protocol construction, synthetic data generation, and adversarial validation into a unified pipeline. The framework translates natural language research questions into structured causal protocols and executable analysis code by first constructing a protocol and then generating synthetic datasets using Structural Causal Models (SCMs) with known ground-truth effects. This synthetic-data step can also support pipeline development when access to confidential data, such as medical data, is restricted. The generated analysis is then evaluated under controlled violations of identification assumptions. We evaluate ARA on the Automated Causal Reasoning Benchmark, assessing recovery of identification strategies, causal quantities, treatment and outcome variables, and consistency between generated code and approved protocol. Protocol construction and adversarial validation did not consistently improve numerical agreement with benchmark estimates compared with standard LLM-based generation. However, they changed the failure mode: instead of silently returning causal estimates, ARA often surfaced protocol concerns, diagnostic failures, incomplete inference, or downgraded non-causal interpretations. These findings suggest that validity-first automated science systems should be evaluated not only by answer accuracy, but also by whether they indicate when causal claims are unwarranted.
\end{abstract}

%%% instructions 
%up to 9 CONTENT pages
%\pages containing acknowledgments, references, checklist, and optional technical appendices do not count as content pages.

\section{Introduction}
The validation of empirical research remains a major bottleneck in AI-assisted scientific discovery. This challenge is particularly acute in causal inference, where valid conclusions depend on untestable identification assumptions and sensitivity to hidden biases \cite{pearl2009causality}. Recent benchmarks suggest that large language models (LLMs) can fail to detect statistical traps such as Simpson’s paradox and selection bias, even when generating syntactically correct code \cite{du2026icecreamdoesntcause}. This causal gap arises because automated assistants often fail to verify critical identification assumptions such as parallel trends in Difference-in-Differences or exclusion restrictions in instrumental variables. Consequently, AI-generated results may be biased due to violations of the assumptions required for valid causal inference.

Early applications of LLMs focused on information retrieval, where models were used to answer questions based on available knowledge \cite{cosme2024systematic}. More recently, LLM-based agents have expanded beyond retrieval and are increasingly used to generate new analytical artifacts \cite{swanson2025virtual, schmidgall2025agent}. This shift is important in empirical research settings, where the full dataset may not be available to the AI system because of privacy, confidentiality, or governance constraints. In such settings, synthetic data can support protocol development, code generation, and validation against known data structures before analysis code is applied to real-world data \cite{giuffre2023harnessing}. 

This challenge is especially relevant in medical and health data science, where causal analyses often rely on sensitive electronic health records, claims data, or registries that cannot be freely shared with external AI systems \cite{feuerriegel2024causal}. Clinical questions are also vulnerable to confounding, selection bias, immortal time bias, informative censoring, outcome misclassification, and measurement error. A test-driven causal framework could support these settings by translating clinical questions and data dictionaries into explicit causal protocols, generating synthetic patient-level failure scenarios, and requiring analysis code to detect assumption violations before real-world deployment. The same framework could support rapid first-pass target trial emulation (TTE) protocol development, producing trial-like specifications with defined eligibility criteria, treatment strategies, time zero, follow-up, outcomes, estimands, and adjustment sets, while preserving the need for clinical judgment and formal statistical review \cite{hernan2022target}.

This motivates our central question: how can LLM-based systems support the specification, validation, and implementation of causal analyses from retrospective data when access to real-world data is constrained and causal assumptions require explicit validation?

Existing solutions to this problem use orchestrated multi-agent systems (MAS) \cite{bazgir2025causalmas}. Current implementations generally fall into three technical categories: (1) collaborative reasoning systems that debate structural assumptions or domain priors \cite{khatibi2025alcm, ban2025integratinglargelanguagemodel}; (2) domain-specific agents designed for fields such as epidemiology \cite{liu2026aiagentepidemiology}; and (3) end-to-end analysis pipelines like Causal-Copilot \cite{wang2025causalcopilot} and CausalAgent \cite{zhu2026causalagent}. While these systems improve upon single-prompt architectures, they generally rely on cooperative refinement or syntax-level self-correction \cite{verma2025causal}. They typically lack mechanisms that stress-test causal assumptions through adversarial scenarios.

To bridge this gap, we propose a test-driven framework for causal inference, in which identification assumptions are converted into adversarial tests prior to analysis. Given a user-provided research question and a data dictionary, the system constructs a research protocol through an adversarial review loop in which specialized agents iteratively propose, critique, and revise methodological assumptions. The finalized protocol is then translated into Structural Causal Models (SCMs), which generate synthetic failure scenarios corresponding to violations of these assumptions \cite{bongers2021foundations}. The generated analysis code must detect and report these violations before execution on real-world data. The framework coordinates this process through structured interaction among specialized agents, including a Principal Investigator (PI), reviewers, and an implementation agent. These agents jointly design and implement the analysis. This design enforces methodological validity as a prerequisite for code execution, converting identification assumptions into executable validation tests.

This adversarial approach prioritizes methodological rigor over simple execution and reduces the risk of automated p-hacking. Our primary contributions include: (1) a test-driven validation framework that operationalizes causal assumptions as executable tests; (2) a structured multi-agent design that decomposes responsibilities across protocol design, validation, and implementation; and (3) an empirical evaluation across standard econometric designs. We evaluate whether this framework improves recovery of causal design specifications, including identification strategies and causal quantities, while supporting systematic validation of downstream analysis code. These results suggest a practical direction for improving the reliability of AI-assisted empirical research.

\section{Method}
\begin{figure}
    \centering
    \includegraphics[width=0.5\linewidth]{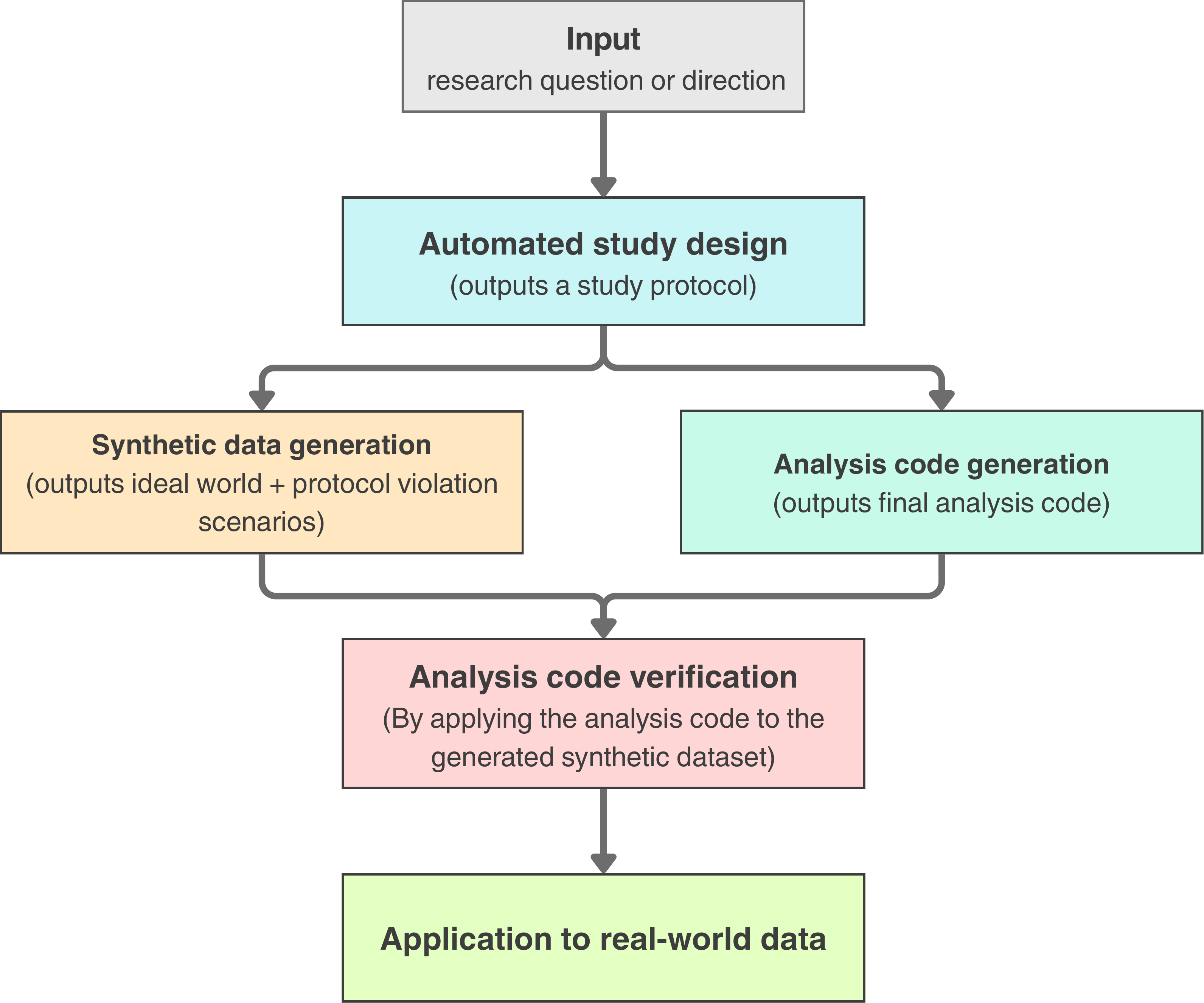}
    \vspace{0.5em} 
    \caption{\textbf{Schematic overview of the multi-agent test-driven causal analysis framework.} The framework comprises three stages: study design, SCM-based synthetic data generation, and code generation with adversarial validation. The approved protocol is converted into a Structural Causal Model (SCM) that produces baseline and failure-scenario datasets. Generated analysis code is iteratively reviewed and validated against these synthetic benchmarks before deployment on real-world data.
    \vspace{0.5em}
}
    \label{fig:architecture}
\end{figure}

We present an end-to-end pipeline for generating, validating, and evaluating causal inference analyses from a free-text research question. The system consists of four sequential phases. In the \textit{Protocol Design Phase}, a multi-agent system constructs and validates a structured study protocol. In the \textit{Code Generation Phase}, two tasks are performed in parallel: the system produces an executable analysis script, and generates synthetic data aligned with the protocol, which is later used in the \textit{data validation phase}. During this phase, the code is validated through multiple quality gates, and evaluated against the synthetic ground truth. 
Finally, the validated script is applied to the real-world research dataset.
The first two phases communicate exclusively through a finalized study protocol and share no additional state. A schematic overview of the system is shown in Figure~\ref{fig:architecture}. In the following subsections, we describe each component of the system in more detail. Additional system specifications are provided in the appendix.

\subsection{Problem formulation}
We consider the problem of automated causal analysis generation from a natural language research question. Given a dataset $D$ and a research query $q$, the goal is to produce an analysis function $f$ that estimates a causal quantity of interest $\tau$, such that $f(D) \rightarrow \hat{\tau}$.

A key challenge is that the validity of $\hat{\tau}$ depends on a set of identification assumptions $\mathcal{A}$ (e.g., parallel trends, exclusion restrictions, continuity). These assumptions are typically untestable from observed data alone and must be externally validated. External knowledge is therefore essential, for example, for specifying plausible causal graphs, identifying potential confounders, and assessing whether the assumptions in $\mathcal{A}$ are credible. In ARA, the protocol-design agent can query external sources and methodological references to ground these decisions, although the present evaluation focuses on benchmark recovery of causal design components rather than independent validation of retrieved evidence.

We define a causal analysis as \textit{valid} if it satisfies two conditions: (i) it produces accurate estimates when the identification assumptions hold, and (ii) it explicitly declares its assumptions and assesses those that can be evaluated from the available data.

\subsection{Test-driven validation framework}
\label{test_driven_val_framework}
To operationalize this notion of validity, we verify the protocol implementation on synthetic data. During this process, the system generates a baseline dataset where causal assumptions hold, and up to three additional datasets with injected violations.

Given a study protocol $\mathcal{P}$, we construct a set of synthetic data-generating processes $\{\mathcal{M}_k\}$ using Structural Causal Models (SCMs). These include both baseline settings, where identification assumptions hold, and failure scenarios, where specific assumptions are violated.

Each dataset $D_k \sim \mathcal{M}_k$ defines a test case with known ground-truth properties. An analysis function is evaluated on each dataset, and its outputs are compared against expected effect behavior and diagnostic outcomes.

This framework converts identification assumptions into executable tests, requiring analysis methods to demonstrate robustness across both valid and adversarial settings.

\subsection{Adversarial protocol construction}
We construct study protocols using a multi-agent adversarial loop. A proposal agent generates candidate designs, while a reviewer agent critiques them by identifying violations of identification assumptions and missing robustness checks.

This interaction can be viewed as an iterative constraint-refinement process that enforces identification assumptions prior to evaluation. The loop terminates when all identified issues are resolved, producing a protocol that defines both the target analysis and its validation requirements.

\subsection{Protocol design}
The protocol design phase is implemented as a two-agent system operating in an iterative loop. A \textit{Principal Investigator} (PI) agent proposes a causal study design given a research question and optionally a data schema. The PI grounds its outputs using external tools, including web search and retrieval over methodological references \cite{Craige080505}, and operates autonomously without user clarification. A second agent, the \textit{Reviewer}, evaluates each proposed protocol by challenging identification assumptions, such as parallel trends or exclusion restrictions, and by verifying the inclusion of appropriate robustness analyses, including placebo tests, sensitivity analyses, and negative controls.

The interaction between the PI and Reviewer forms an adversarial review loop. In each iteration, the PI proposes or revises a protocol, and the Reviewer critiques it by identifying methodological weaknesses. The Reviewer produces a structured decision indicating approval or the need for revision. Approval is only permitted when all identified issues are resolved, and at least two full review rounds are required before termination. The loop is bounded by a maximum number of iterations to ensure tractability.

The interaction is managed through a state-based controller with conditional routing between agents and tool calls. To control context length, the system retains the original task and a limited window of recent messages, while constraining tool usage to prevent excessive or redundant queries.

Upon approval, the PI emits a finalized study protocol that specifies the research question, estimand, study design, identification strategy, and required diagnostics. This protocol and the data schema are passed to the second phase.

\subsection{Code pipeline}
The second phase implements the test-driven validation framework defined in Section~\ref{test_driven_val_framework} through synthetic data generation and iterative code validation.

\subsubsection{Synthetic data generation}
The pipeline begins by constructing a synthetic evaluation environment aligned with the protocol. A planning agent identifies a set of failure scenarios corresponding to violations of key identification assumptions, such as violations of parallel trends, manipulation at a regression discontinuity cutoff, or instrument invalidity. A second planning step produces a detailed data-generating process, including structural equations, a direct acyclic graph, parameter ranges, and dataset structure. 
Datasets are generated to match the schema of the real data (column names and types) but are otherwise constructed as functional stress-tests of the identification strategy rather than calibrated reproductions of empirical variance.
A code-generation agent then implements a data generator capable of producing datasets under both effect-present and null conditions. The generated data is validated against structural constraints, including sample size, clustering, and temporal structure. The validation check determines whether the analysis recovers the injected ground-truth effect and whether its diagnostics correctly detect the planted violations.

\subsubsection{Analysis code generation}
Once the data is constructed, the pipeline moves on to the analysis code generation. The free-text protocol is converted into a structured specification using a schema-based extractor. This representation includes the study design, estimand, population, covariates, model specification, identification assumptions, and robustness requirements. The structured protocol serves as the authoritative specification for all downstream agents.

An architecture planning agent then produces an implementation plan that defines the analysis workflow, required components, statistical methods, and expected outputs. All method choices are verified through external search before being fixed.

A code-generation agent produces a complete Python analysis script based on the architecture plan and data schema. To prevent leakage of evaluation information, the agent is explicitly restricted from accessing ground-truth parameters from the synthetic data generator. To ensure robustness under token constraints, the agent first generates an executable skeleton and incrementally fills in missing components.

The generated script is refined through an iterative validation loop consisting of multiple quality gates. The script is first checked for syntactic correctness, followed by smoke tests that verify basic execution and output structure. It is then evaluated by an auditing agent that checks consistency with the structured protocol and identifies methodological or implementation issues. After passing the audit, the script is executed on synthetic data, and its outputs are compared against ground truth. Any failures at any stage trigger targeted fixes by a dedicated fixing agent, and the process repeats until convergence or a maximum number of iterations is reached. The final output is the terminal validation-loop script, together with a complete audit trail. This script is either an approved script that passed the synthetic validation criteria or, when approval is not reached, the last script produced within the maximum iteration budget.

\subsection{Engineering safeguards}
The system incorporates several safeguards to ensure robustness and validity. 
\begin{itemize}
    \item Ground-truth information is isolated from all code-generating agents to prevent leakage. 
    \item Statistical methods are verified through external sources before implementation. 
    \item Diagnostic procedures are constrained to be consistent with the chosen study design. 
    \item Computationally intensive procedures are parallelized to ensure tractability. 
    \item Token usage is monitored and controlled to maintain efficiency, and all intermediate states are version-controlled to ensure reproducibility.
\end{itemize}

\section{Experiments}
We evaluated our pipeline on the Automated Causal Reasoning Benchmark \cite{causalreasoningbenchmark2026}, a dataset of observational studies paired with real tabular data and a data dictionary. Each case includes a full causal question (precise estimand, strategy, and variable names) and a vague question (underspecified). We run both variants for each case, yielding two conditions per benchmark item.

We selected 33 benchmark cases using a stratified round-robin procedure based on the benchmark's identification strategy labels. First, cases within each identification strategy were randomly shuffled. The shuffled strategy-specific groups were then interleaved in round-robin order by taking one case from each strategy at a time. We used the first 33 cases from this interleaved list for the experiments. This procedure was intended to reduce overrepresentation of strategies that appear earlier or more frequently in the dataset.

For each benchmark case, we evaluated two question formulations with different levels of specificity: an explicit causal question and a vague version of the same question. In the vague formulation, identifying hints were removed, making the causal task less directly specified. This setup allowed us to test the robustness of the pipeline to underspecified user queries. 

\paragraph{Pipeline.} For each run, the code pipeline was permitted up to 12 iterations. Within each iteration, the fixer was allowed up to 5 repair attempts before the pipeline regenerated the script from scratch. We tracked three script checkpoints: the initial generated script (pre-audit), the post-audit version, and the final validation-loop checkpoint.

\paragraph{Baseline.} We also ran a vanilla baseline for each question formulation. In this setting, the original question text and data dictionary were provided directly to Claude Opus 4.6, with only minimal formatting changes, and the model was prompted to generate a single analysis script. The resulting script was executed on the corresponding dataset, without protocol generation, audit, or iterative refinement.

\paragraph{Evaluation.} We evaluated each run according to script generation success, execution success, recovered identification strategy, extracted causal quantity, selected treatment and outcome variables, and reported effect estimates and standard errors. For the pipeline condition, we additionally report audit pass rates and the fraction of cases reaching each checkpoint.

\paragraph{Code Availability.} The code, prompts, configuration files, and reproduction scripts needed to reproduce the main experiments are available in an anonymized repository created using Anonymous 4open.science: \url{https://anonymous.4open.science/r/ARA-08DB}. 

\section{Results and Discussion}
We evaluated ARA along two complementary dimensions. First, we assessed whether the generated analyses recovered the correct causal problem formulation, including the treatment, outcome, causal quantity, and identification strategy. Second, we assessed whether the resulting statistical outputs agreed with the benchmark estimates, including the direction of the confidence interval relative to zero and numerical agreement with benchmark effects. We begin by defining the pipeline checkpoints used in these comparisons.

\subsection{Methods evaluated}

ARA outputs are interpreted by checkpoint because the validation loop can terminate in two ways: by satisfying the synthetic validation criteria, or by reaching the 12-iteration budget. We therefore report three checkpoints: the initial script before audit, the post-audit script, and the final checkpoint, defined as the terminal output of the validation loop. The final checkpoint is not necessarily a fully approved script.

This distinction is important for the remaining results. The audit stage enforces protocol consistency, while the validation loop tests recovery of known synthetic effects and detection of planted violations. In some runs, the auditor introduced or enforced protocol constraints that improved methodological compliance but made convergence to a benchmark-like estimate harder within the fixed iteration budget. Thus, final-checkpoint outputs should be interpreted as the terminal state of the validation process, not as uniformly validated analyses. We return to technical and inferential completion failures in Section~\ref{sec:fails}.

\subsection{Directional agreement and conservative CI behavior}
\begin{figure}
    \centering
    \includegraphics[width=0.7\linewidth]{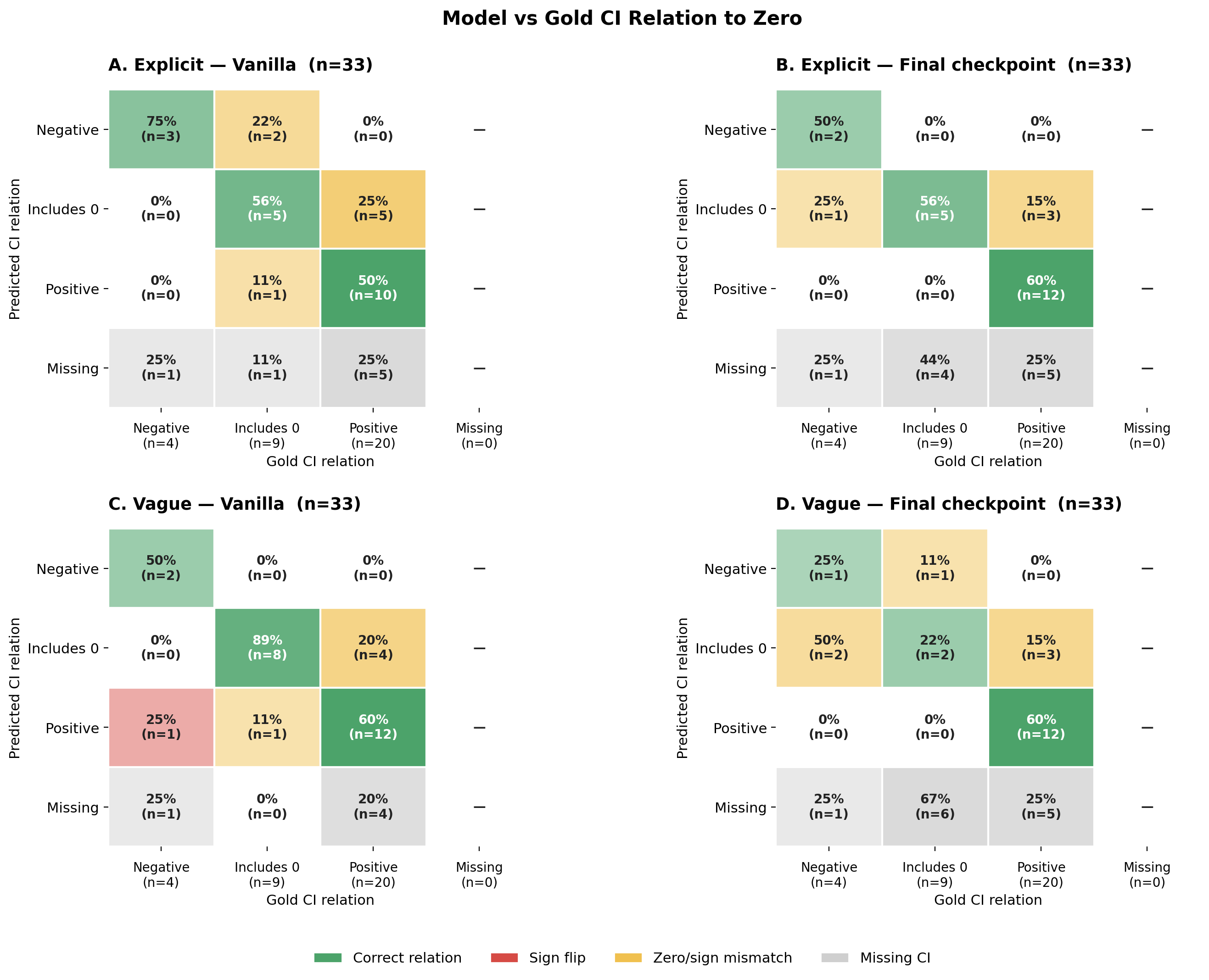}
    \caption{\textbf{Directional agreement between generated and benchmark confidence intervals.}
Rows show the generated confidence-interval relation to zero, and columns show the benchmark relation. Panels compare vanilla generation and the ARA final checkpoint for explicit and vague prompts. Green cells indicate matching direction, yellow cells indicate zero/sign mismatches, red cells indicate sign flips, and gray cells indicate missing confidence intervals. “Final checkpoint” denotes the terminal output of the validation loop, either an approved script or the last script produced within the 12-iteration budget.}
    \label{fig:ci_confusion}
\end{figure}

We evaluated whether each generated confidence interval fell on the same side of zero as the benchmark interval, included zero, or was missing (Figure~\ref{fig:ci_confusion}). This comparison is safety-relevant because a sign flip can reverse the interpretation of benefit and harm. In this 33 case sample, across explicit and vague prompts, ARA produced no sign flips at the final checkpoint (0/33 in both). By contrast, the vanilla baseline produced one sign flip in the vague condition (1/33), reporting a positive interval when the benchmark interval was negative.

ARA's errors were more often conservative or incomplete rather than directionally inverted. In the explicit condition, ARA returned intervals that included zero for 4/33 cases in which the benchmark interval was strictly positive or negative; in the vague condition, this occurred in 5/33 cases. These cases reduce strict benchmark agreement, but they are qualitatively different from sign flips: they indicate weaker support for a directional claim rather than evidence in the opposite direction. ARA also produced more missing confidence intervals than vanilla. In explicit prompts, missing intervals occurred in 11/33 ARA final-checkpoint outputs versus 7/33 vanilla outputs. In vague prompts, missing intervals occurred in 13/33 ARA outputs versus 5/33 vanilla outputs. These missing intervals are penalized in aggregate benchmark metrics, but they should not be interpreted uniformly: some reflect execution or reporting failures, while others reflect validity-driven withholding of a causal interval.

\subsection{Causal specification recovery}
\begin{figure}
    \centering
    \includegraphics[width=0.7\linewidth]{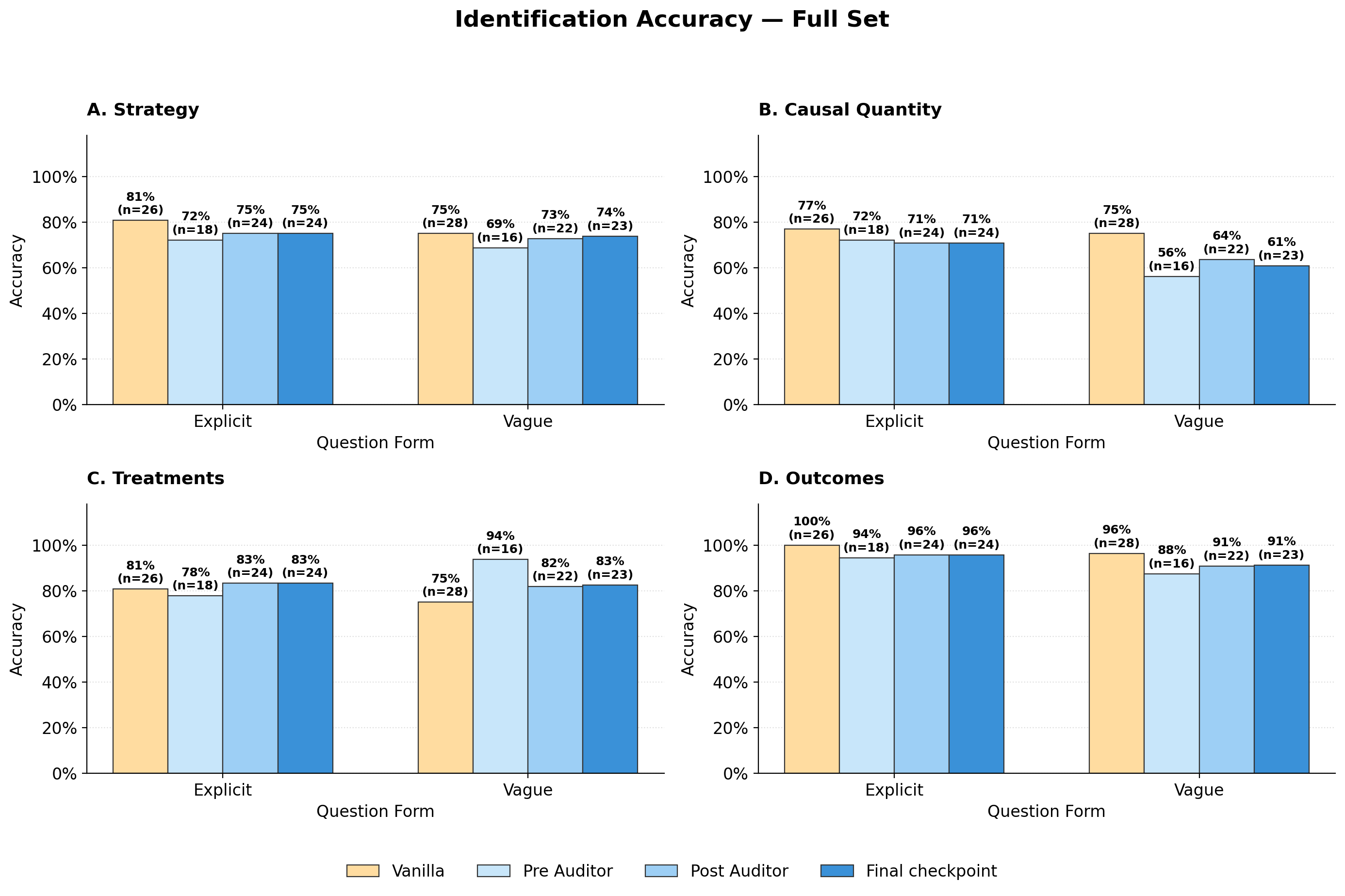}
    \caption{\textbf{Causal specification recovery across prompt types and pipeline checkpoints.} Panels show recovery of the benchmark identification strategy, causal quantity, treatment variable, and outcome variable for explicit and vague prompts. ARA is evaluated at the pre-audit, post-audit, and final-checkpoint stages; vanilla is evaluated as a direct single-script baseline. Bars report the fraction of evaluable outputs that matched the benchmark label. Denominators vary across bars because failed or non-evaluable outputs are excluded from the corresponding metric.}
    \label{fig:identification}
\end{figure}

We next evaluated recovery of the benchmark treatment, outcome, identification strategy, and causal quantity (Figure~\ref{fig:identification}). Outcome recovery was high across systems and was not the main bottleneck. Treatment recovery showed the clearest benefit from ARA under vague prompts: the final checkpoint recovered the treatment in 83\% of evaluable vague cases, compared with 75\% for vanilla. Strategy recovery improved from ARA's pre-audit checkpoint to the final checkpoint, but remained similar to the vanilla baseline. Causal quantity recovery was the weakest component; ARA improved after auditing in the vague condition, but still lagged behind vanilla on this strict benchmark label.

These results suggest that protocol construction and auditing can help recover parts of the causal specification, especially when the user query is underspecified. However, they do not guarantee better benchmark-label recovery across all components. ARA's advantage is therefore not uniform specification accuracy, but a more explicit and reviewable path from question to protocol to analysis.

\subsection{Statistical agreement with benchmark outputs}
\begin{figure}
    \centering
    \includegraphics[width=0.95\linewidth]{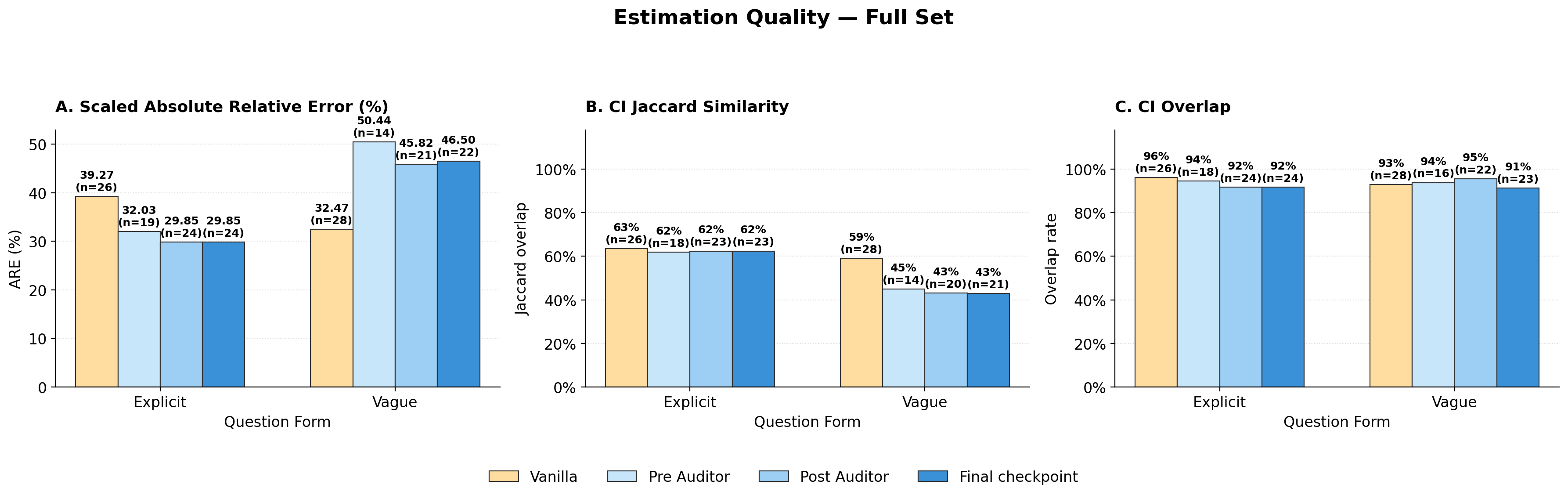}
    \caption{\textbf{Statistical agreement with benchmark estimates.} Panels report scaled absolute relative error, confidence-interval Jaccard similarity, and binary confidence-interval overlap for explicit and vague prompts. Scaled absolute relative error measures point-estimate agreement, CI Jaccard measures interval similarity, and CI overlap indicates whether the generated and benchmark intervals overlap at all. These metrics evaluate numerical benchmark reproduction and should be interpreted together with the directional agreement matrices in Figure 2.}
    \label{fig:estimation}
\end{figure}

We evaluated numerical agreement using scaled absolute relative error, confidence-interval Jaccard similarity, and binary confidence-interval overlap (Figure~\ref{fig:estimation}). In explicit questions, ARA improved point-estimate agreement relative to vanilla: scaled absolute relative error decreased from 39.27\% for vanilla to 29.85\% at the ARA final checkpoint. In vague questions, the pattern reversed: vanilla achieved lower error than ARA, and vanilla also had higher CI Jaccard similarity. This may be explained by ARA's tendency to produce wider CIs which can be interpreted as a form of conservative approach. Binary CI overlap was high for both systems across prompt types.

These metrics measure benchmark reproduction, not causal safety. They favor systems that return numerical estimates and intervals. The contrast between Figure~\ref{fig:ci_confusion} and Figure~\ref{fig:estimation} is therefore important: vanilla often looks stronger under strict numerical agreement, while ARA more often returns intervals that include zero, lacks standard errors or confidence intervals in some runs, or raises protocol-level validity concerns. This supports the central interpretation that adversarial validation trades off answer production against causal-validity screening.

\subsection{Detection of assumption violations and causal downgrading}
We define a detected causal-validity concern as a case in which the pipeline flags a violation of an identification assumption, fails a required diagnostic, downgrades a causal claim to descriptive, or states that no causal identification is possible.

These concerns were most useful for interpreting missing or non-causal outputs: several ARA failures reflected explicit recognition that required instruments, cutoffs, compliance measures, negative controls, or covariates were unavailable, rather than only script execution failure.

\subsection{Failure modes}
\label{sec:fails}
Failure rates decreased across ARA checkpoints, although final-checkpoint ARA did not outperform vanilla on raw completion. In explicit prompts, hard script failures decreased from 14/33 before audit to 9/33 at the final checkpoint, and incomplete inference decreased from 15/33 to 10/33. In vague prompts, hard script failures decreased from 19/33 to 11/33, and incomplete inference decreased from 19/33 to 12/33. By comparison, vanilla had 7/33 hard script or incomplete-inference failures in explicit prompts and 5/33 in vague prompts. Thus, auditing and validation improved ARA's internal completion profile, but the simpler baseline more often produced complete executable outputs.

Missing confidence intervals were heterogeneous. Some reflected hard script failures, where no usable analysis summary was produced. Others reflected incomplete inference, where the script produced partial output but omitted the standard error or confidence interval required for benchmark comparison. A third group reflected causal-validity concerns raised by ARA, such as unavailable instruments or cutoffs, missing exposure or compliance information, absent negative controls, missing covariates, or downgrading a causal estimand to a descriptive association.

This distinction matters because a missing CI is not always the same failure. In the aggregate benchmark comparison, all missing CIs count against the system. Methodologically, however, a missing interval caused by a validity concern is different from a technical execution failure. The validity flags themselves are model-generated and require manual review, but they show that ARA can surface the kinds of assumptions and data limitations that direct answer-generation baselines may leave implicit.

More broadly, these findings suggest that reliability gains may come less from prompting a model to be careful and more from encoding domain-specific best practices, assumptions, and constraints into the artifacts the model must produce and satisfy, such as protocols, diagnostics, and validation tests. Automated causal-analysis systems should therefore be evaluated not only by numerical agreement with a benchmark estimate, but also by whether they avoid unsupported directional claims.

\subsection{Limitations}
The evaluation includes only 33 benchmark cases, so small counts, including the observed sign flip, should be interpreted cautiously. The final checkpoint is not always a formally approved script; it is the terminal output of the validation loop, either approved or stopped at the 12-iteration budget. Longer validation budgets may reduce technical and inferential failures, but could increase cost and may not eliminate protocol-level non-identification. The interaction between auditing and validation also requires further study: stricter auditing can improve protocol compliance while reducing short-run estimation convergence.

The validity flags produced by ARA are not ground-truth labels. They identify candidate concerns about assumptions, diagnostics, or data availability, but require manual review by domain and methods experts. The results are also model-dependent because both the protocol and code-generation stages rely on specific LLMs and agent prompts. Finally, the pipeline is substantially more complex than direct script generation, which increases engineering burden and creates additional failure points. Future work should separate approved and non-approved terminal outputs and evaluate robustness across model families, prompts, and validation budgets.

\section{Conclusion}
ARA is a test-driven framework for turning causal research questions into structured protocols, synthetic validation tests, and executable analysis code. It improved selected specification and directional-safety outcomes in this benchmark, including treatment recovery under vague prompts and no observed sign flips at the final checkpoint. Its main contribution is a change in the error profile of automated causal analysis: some unsupported causal claims, incomplete inferential outputs, and protocol-level validity concerns become visible rather than remaining implicit behind a numerical estimate. This safety-oriented behavior did not consistently translate into better numerical benchmark reproduction relative to a direct baseline.
For validity-first automated science systems, this suggests that evaluation should measure not only whether a system returns the benchmark answer, but also whether it recognizes when a causal answer is not yet warranted.

\FloatBarrier

\begin{ack}
Funding: 

Competing interests:

Use unnumbered first level headings for the acknowledgments. All acknowledgments
go at the end of the paper before the list of references. Moreover, you are required to declare
funding (financial activities supporting the submitted work) and competing interests (related financial activities outside the submitted work).
More information about this disclosure can be found at: \url{https://neurips.cc/Conferences/2026/PaperInformation/FundingDisclosure}.

Do {\bf not} include this section in the anonymized submission, only in the final paper. You can use the \texttt{ack} environment provided in the style file to automatically hide this section in the anonymized submission.
\end{ack}

\bibliographystyle{unsrtnat}
\bibliography{references}

%%%%%%%%%%%%%%%%%%%%%%%%%%%%%%%%%%%%%%%%%%%%%%%%%%%%%%%%%%%%

\appendix

\section{Technical appendices and supplementary material}

\section*{System Specifications}

% ---------------------------------------------------------------
\subsection*{Design Pipeline}
% ---------------------------------------------------------------

The study protocol is produced by a two-agent LangGraph workflow in which a
Principal Investigator (PI) agent and a Reviewer agent iterate until the
Reviewer issues approval.

\begin{table}[H]
\centering
\begin{tabular}{ll}
\hline
\textbf{Parameter} & \textbf{Value} \\
\hline
Orchestration framework & LangGraph (\texttt{StateGraph}) \\
PI model - initial pass & \texttt{gpt-5.2} via Azure OpenAI \\
PI model - revision passes & \texttt{gpt-5.2-chat} via Azure OpenAI \\
Reviewer model & \texttt{gpt-5.2-chat} via Azure OpenAI \\
Azure OpenAI API version & \texttt{2024-02-01} \\
Web search tool & Tavily (\texttt{max\_results=10}, domain blacklist applied) \\
RAG retriever & Chroma vectorstore; \texttt{text-embedding-3-small} embeddings \\
RAG chunking & chunk size 1{,}000 tokens, overlap 200 tokens \\
Context window (PI node) & last 15 messages \\
Retry policy & up to 4 attempts; delays 5, 15, 30, 60\,s \\
\hline
\end{tabular}
\end{table}

\noindent The PI agent has access to two tools: \texttt{web\_search} (Tavily) and
\texttt{reference\_paper\_search} (retriever over the reference PDF).
Model switching is dynamic: \texttt{gpt-5.2} is used for the first PI pass;
all subsequent PI and Reviewer calls use \texttt{gpt-5.2-chat}.

% ---------------------------------------------------------------
\subsection*{Code Generation Pipeline}
% ---------------------------------------------------------------

Given the finalised protocol, a five-agent pipeline generates, audits, and
iteratively repairs analysis code against a synthetic ground-truth contract.

\begin{table}[H]
\centering
\begin{tabular}{ll}
\hline
\textbf{Parameter} & \textbf{Value} \\
\hline
Model & \texttt{claude-opus-4-6} \\
API provider & Anthropic via Azure AI Foundry \\
Anthropic API version & \texttt{2023-06-01} \\
Beta features & \texttt{prompt-caching-2024-07-31} \\
Temperature & \texttt{0.0} (deterministic) / \texttt{1.0} when thinking enabled \\
Extended thinking & Enabled, budget 8{,}000 tokens per call \\
Max output tokens per call & 16{,}000 \\
Fallback model & \texttt{gpt-5.2} (OpenAI-compatible endpoint) \\
Max refinement iterations & 12 \\
Retry policy & up to 5 attempts; delays 5, 15, 30, 60\,s \\
\hline
\end{tabular}
\end{table}

\noindent The five agents - \texttt{spec\_llm}, \texttt{planner\_llm},
\texttt{coder\_llm}, \texttt{auditor\_llm}, and \texttt{fixer\_llm} - share
the same model and configuration. An optional Tavily web-search tool is
available to the coder and fixer agents during code generation.

% ---------------------------------------------------------------
\subsection*{Analysis Libraries}
% ---------------------------------------------------------------

The generated analysis script (\texttt{analysis.py}) depends on the following
Python packages:

\begin{itemize}
  \item \texttt{numpy}
  \item \texttt{pandas}
  \item \texttt{statsmodels}
  \item \texttt{linearmodels} (IV2SLS, Kleibergen--Paap $F$-statistic)
  \item \texttt{scipy}
\end{itemize}

% ---------------------------------------------------------------
\subsection*{Execution Environment}
% ---------------------------------------------------------------

\begin{table}[H]
\centering
\begin{tabular}{ll}
\hline
\textbf{Item} & \textbf{Value} \\
\hline
Operating system & Ubuntu 22.04.5 LTS (Jammy Jellyfish) \\
Kernel & Linux 5.15.0-164-generic \\
Architecture & x86\_64 \\
CPU cores & 40 \\
RAM & 251\,GB \\
Python & 3.10.12 \\
\hline
\end{tabular}
\end{table}

\subsection*{Failure audit and missing-output taxonomy}

\begin{table}[H]
\centering
\caption{Failure audit for pipeline and baseline runs. The table reports hard script failures, incomplete inference, and numeric failure rates by run type and checkpoint. The final checkpoint denotes the terminal validation-loop output and is not necessarily a formally approved script.}
\label{tab:technical-failure-rates}
\begin{tabular}{llrrrr}
\toprule
Condition & Stage & $N$ & Hard failures & Incomplete inference & Any numeric failure \\
\midrule
Vanilla explicit & Final checkpoint & 33 & 7 (21.2\%) & 7 (21.2\%) & 7 (21.2\%) \\
Vanilla vague & Final checkpoint & 33 & 5 (15.2\%) & 5 (15.2\%) & 5 (15.2\%) \\
ARA explicit & Pre-audit & 33 & 14 (42.4\%) & 15 (45.5\%) & 15 (45.5\%) \\
ARA explicit & Audited & 33 & 9 (27.3\%) & 10 (30.3\%) & 10 (30.3\%) \\
ARA explicit & Final checkpoint & 33 & 9 (27.3\%) & 10 (30.3\%) & 10 (30.3\%) \\
ARA vague & Pre-audit & 33 & 19 (57.6\%) & 19 (57.6\%) & 19 (57.6\%) \\
ARA vague & Audited & 33 & 12 (36.4\%) & 13 (39.4\%) & 13 (39.4\%) \\
ARA vague & Final checkpoint & 33 & 11 (33.3\%) & 12 (36.4\%) & 12 (36.4\%) \\
\bottomrule
\end{tabular}
\end{table}

\subsection*{Model pricing and cost assumptions}

\begin{table}[H]
\centering
\caption{Estimated API cost by generation condition.}
\label{tab:pricing-summary}
\begin{tabular}{lrrrr}
\toprule
Condition & Runs & Calls & Total cost & Mean cost/run \\
\midrule
Vanilla explicit & 33 & 66 & \$4.84 & \$0.15 \\
Vanilla vague & 33 & 66 & \$4.56 & \$0.14 \\
ARA explicit total & 32* & 2034 & \$410.49 & \$12.83 \\
ARA vague total & 33 & 1744 & \$394.14 & \$11.94 \\
\bottomrule
\end{tabular}
\end{table}

* One case was not executed due to content management policy triggering.

%%%%%%%%%%%%%%%%%%%%%%%%%%%%%%%%%%%%%%%%%%%%%%%%%%%%%%%%%%%%

\makeatletter
\if@preprint\else
  \newpage
  \section*{NeurIPS Paper Checklist}

\begin{enumerate}

\item {\bf Claims}
    \item[] Question: Do the main claims made in the abstract and introduction accurately reflect the paper's contributions and scope?
    \item[] Answer: \answerYes{} % Replace by \answerYes{}, \answerNo{}, or \answerNA{}.
    \item[] Justification: The abstract and introduction accurately describe the paper’s scope: ARA is presented as a framework for protocol construction, synthetic data generation, and adversarial validation, and the paper explicitly states that the main benefit is changing failure modes rather than uniformly improving numerical benchmark agreement.

\item {\bf Limitations}
    \item[] Question: Does the paper discuss the limitations of the work performed by the authors?
    \item[] Answer: \answerYes{} % Replace by \answerYes{}, \answerNo{}, or \answerNA{}.
    \item[] Justification: The paper includes a dedicated Limitations section. It discusses the small benchmark size, the fact that the final checkpoint is not always formally approved, model dependence, validation-budget constraints, and added engineering complexity.

\item {\bf Theory assumptions and proofs}
    \item[] Question: For each theoretical result, does the paper provide the full set of assumptions and a complete (and correct) proof?
    \item[] Answer: \answerNA{} % Replace by \answerYes{}, \answerNo{}, or \answerNA{}.
    \item[] Justification: The paper does not present formal theoretical results, theorems, or proofs. It defines a validation framework and evaluates it empirically.

    \item {\bf Experimental result reproducibility}
    \item[] Question: Does the paper fully disclose all the information needed to reproduce the main experimental results of the paper to the extent that it affects the main claims and/or conclusions of the paper (regardless of whether the code and data are provided or not)?
    \item[] Answer: \answerYes{}{} % Replace by \answerYes{}, \answerNo{}, or \answerNA{}.
    \item[] Justification: The paper discloses the benchmark, sampling procedure, prompt conditions, pipeline checkpoints, iteration budgets, baseline design, evaluation metrics, model configurations, software dependencies, and execution environment. These details provide a reasonable path to reproducing the main experimental results, although exact reproducibility may depend on access to the specified closed-source models and APIs.

\item {\bf Open access to data and code}
    \item[] Question: Does the paper provide open access to the data and code, with sufficient instructions to faithfully reproduce the main experimental results, as described in supplemental material?
    \item[] Answer: \answerYes{} % Replace by \answerYes{}, \answerNo{}, or \answerNA{}.
    \item[] Justification: The code, prompts, configuration files, and reproduction scripts needed to reproduce the main experimental results are provided in an anonymized repository created using Anonymous 4open.science.

\item {\bf Experimental setting/details}
    \item[] Question: Does the paper specify all the training and test details (e.g., data splits, hyperparameters, how they were chosen, type of optimizer) necessary to understand the results?
    \item[] Answer: \answerYes{} % Replace by \answerYes{}, \answerNo{}, or \answerNA{}.
    \item[] Justification: The paper specifies the benchmark, number of cases, stratified round-robin sampling procedure, explicit and vague prompt conditions, maximum refinement iterations, repair attempts, baseline setup, checkpoints, and evaluation metrics. Additional implementation details are provided in the technical appendix.

\item {\bf Experiment statistical significance}
    \item[] Question: Does the paper report error bars suitably and correctly defined or other appropriate information about the statistical significance of the experiments?
    \item[] Answer: \answerNo{} % Replace by \answerYes{}, \answerNo{}, or \answerNA{}.
    \item[] Justification: The paper reports descriptive metrics, counts, proportions, confidence-interval agreement, and failure rates, but it does not report error bars, statistical significance tests, or uncertainty intervals for comparisons across systems. 

\item {\bf Experiments compute resources}
    \item[] Question: For each experiment, does the paper provide sufficient information on the computer resources (type of compute workers, memory, time of execution) needed to reproduce the experiments?
    \item[] Answer: \answerYes{} % Replace by \answerYes{}, \answerNo{}, or \answerNA{}.
    \item[] Justification: The appendix reports the operating system, kernel, CPU architecture, number of CPU cores, RAM, Python version, model/API settings, iteration budgets, and estimated API costs.
    
\item {\bf Code of ethics}
    \item[] Question: Does the research conducted in the paper conform, in every respect, with the NeurIPS Code of Ethics \url{https://neurips.cc/public/EthicsGuidelines}?
    \item[] Answer: \answerYes{}{} % Replace by \answerYes{}, \answerNo{}, or \answerNA{}.
    \item[] Justification: The research conforms to the NeurIPS Code of Ethics. The study evaluates an automated causal-analysis framework using benchmark observational-study cases and synthetic validation scenarios, and it is designed to reduce unsupported causal claims by making assumptions, diagnostics, and validity concerns explicit.

\item {\bf Broader impacts}
    \item[] Question: Does the paper discuss both potential positive societal impacts and negative societal impacts of the work performed?
    \item[] Answer: \answerYes{} % Replace by \answerYes{}, \answerNo{}, or \answerNA{}.
    \item[] Justification: The paper discusses possible benefits for medical and health data science and emphasizes safeguards against unsupported causal claims. 
    
\item {\bf Safeguards}
    \item[] Question: Does the paper describe safeguards that have been put in place for responsible release of data or models that have a high risk for misuse (e.g., pre-trained language models, image generators, or scraped datasets)?
    \item[] Answer: \answerNA{}{} % Replace by \answerYes{}, \answerNo{}, or \answerNA{}.
    \item[] Justification: The paper does not release a pretrained model, scraped dataset, image generator, or other asset with a high risk for misuse. The released assets consist of code, prompts, configuration files, and reproduction scripts for the ARA evaluation pipeline.

\item {\bf Licenses for existing assets}
    \item[] Question: Are the creators or original owners of assets (e.g., code, data, models), used in the paper, properly credited and are the license and terms of use explicitly mentioned and properly respected?
    \item[] Answer: \answerYes{} % Replace by \answerYes{}, \answerNo{}, or \answerNA{}.
    \item[] Justification: The paper cites the existing benchmark, methodological references, model/API providers, orchestration tools, retrieval tools, vector stores, and Python libraries used in the work.

\item {\bf New assets}
    \item[] Question: Are new assets introduced in the paper well documented and is the documentation provided alongside the assets?
    \item[] Answer: \answerYes{} % Replace by \answerYes{}, \answerNo{}, or \answerNA{}.
    \item[] Justification: The paper introduces and releases new research assets, including the ARA pipeline code, prompts, configuration files, reproduction scripts, and evaluation scripts. These assets are provided through an anonymized repository and documented with setup instructions, usage guidance, limitations, and licensing information.

\item {\bf Crowdsourcing and research with human subjects}
    \item[] Question: For crowdsourcing experiments and research with human subjects, does the paper include the full text of instructions given to participants and screenshots, if applicable, as well as details about compensation (if any)? 
    \item[] Answer: \answerNA{} % Replace by \answerYes{}, \answerNo{}, or \answerNA{}.
    \item[] Justification: The paper does not involve crowdsourcing experiments or direct research with human subjects. It evaluates an automated causal-analysis framework on benchmark observational-study cases and synthetic validation data.

\item {\bf Institutional review board (IRB) approvals or equivalent for research with human subjects}
    \item[] Question: Does the paper describe potential risks incurred by study participants, whether such risks were disclosed to the subjects, and whether Institutional Review Board (IRB) approvals (or an equivalent approval/review based on the requirements of your country or institution) were obtained?
    \item[] Answer: \answerNA{} % Replace by \answerYes{}, \answerNo{}, or \answerNA{}.
    \item[] Justification: The paper does not involve direct interaction with human participants, crowdsourcing, or collection of new human-subject data. Therefore, IRB approval or equivalent review does not appear applicable based on the current study design.

\item {\bf Declaration of LLM usage}
    \item[] Question: Does the paper describe the usage of LLMs if it is an important, original, or non-standard component of the core methods in this research? Note that if the LLM is used only for writing, editing, or formatting purposes and does \emph{not} impact the core methodology, scientific rigor, or originality of the research, declaration is not required.
    %this research? 
    \item[] Answer: \answerYes{} % Replace by \answerYes{}, \answerNo{}, or \answerNA{}.
    \item[] Justification: LLMs are a core component of the method. The paper describes the use of LLM-based agents for protocol design, review, code generation, auditing, and fixing, and the appendix reports the model families, providers, API versions, temperature settings, and related configuration details.

\end{enumerate}%
\fi
\makeatother

\end{document}